# Navigating Process Mining: A Case study using pm4py

LÁSZLÓ KOVÁCS
University of Miskolc, Hungary
Institute of Information Technology
kovacs@iit.uni-miskolc.hu

ALI JLIDI
University of Miskolc, Hungary
Institute of Information Technology
jlidi.ali@student.uni-miskolc.hu

**Abstract:** Process-mining techniques have emerged as powerful tools for analyzing event data to gain insights into business processes. In this paper, we present a comprehensive analysis of road traffic fine management processes using the pm4py library in Python. We start by importing an event log dataset and explore its characteristics, including the distribution of activities and process variants. Through filtering and statistical analysis, we uncover key patterns and variations in the process executions.

Subsequently, we apply various process-mining algorithms, including the Alpha Miner, Inductive Miner, and Heuristic Miner, to discover process models from the event log data. We visualize the discovered models to understand the workflow structures and dependencies within the process. Additionally, we discuss the strengths and limitations of each mining approach in capturing the underlying process dynamics.

Our findings shed light on the efficiency and effectiveness of road traffic fine management processes, providing valuable insights for process optimization and decision-making. This study demonstrates the utility of pm4py in facilitating process mining tasks and its potential for analyzing real-world business processes.

## 1. Introduction

In today's data-driven world, organizations across various industries face the challenge of managing and optimizing complex business processes [1] to enhance efficiency, reduce costs, and improve customer satisfaction. One powerful approach to tackle this challenge is process mining, a discipline that leverages event data to analyze and understand the dynamics of business processes. By uncovering hidden patterns [2], deviations, and bottlenecks in process executions, process mining provides valuable insights for process optimization [3] and decision-making.

The evolution of process mining can be traced back to the need for efficient pattern matching algorithms in different fields such as manufacturing, healthcare, finance, and telecommunications. These algorithms were developed to analyze event logs and discover underlying process models from raw data [4]. Over time, various process mining techniques and algorithms have been proposed to address different aspects of process analysis, ranging from discovering control flow to identifying performance metrics and resource utilization.

Among the notable process mining algorithms are the Alpha Miner, Inductive Miner, and Heuristic Miner, each offering unique strengths and capabilities in extracting process knowledge from event data [5]. The Alpha Miner, for instance, focuses on discovering causal dependencies between activities in a process [6], while the Inductive Miner employs a recursive approach to construct process models based on event logs [7]. On the other hand, the Heuristic Miner utilizes heuristics to infer process models from event data [8], making it suitable for handling noisy or incomplete logs.

In this paper, we discover process mining using the pm4py library [9], a versatile toolkit for process mining tasks in Python. While pm4py provides convenient access to a range of process mining functionalities, our focus extends beyond mere application; we aim to unravel the inner workings of process mining algorithms implemented within pm4py. By understanding the principles and methodologies behind these algorithms [10], we gain deeper insights into their behavior and performance on real-world dataset.

Up to this point, we are working on an open source dataset of road traffic fine management process. [11] Before extracting processes, we will analyze the data trying to understand it. Here comes the power of process mining library because it provides tools and metrics [12] to explore the logs. Once the data is discovered, it become easier to judge the results

In summary, this paper serves as a comprehensive exploration of process mining methodologies, algorithms, and tools, with a particular emphasis on understanding the underlying mechanisms of process discovery. Through our investigation, we aim to provide practitioners and researchers with valuable insights into the theory and practice of process mining, facilitating informed decision-making and innovation in process-centric domains [13].

## 2. Background survey

### 2.1. Overview of process mining

Process mining is an expanding field at the intersection of data mining and business process management. [14] It involves the analysis of event data generated during the execution of business processes to gain insights into process flows, performance, and compliance. The overarching goal of process mining is to enhance process efficiency, discover bottlenecks, and identify opportunities for improvement. [15] Through techniques such as process discovery, conformance checking, and process enhancement, process mining enables organizations to uncover hidden patterns, deviations, and inefficiencies within their operations. By leveraging event logs from information systems, process mining provides a comprehensive overview of how processes are executed in reality, offering valuable insights for optimizing workflows and achieving operational excellence. [16]

Process mining is recognized as a valuable resource within the expansive landscape of Business Process Management (BPM). It can also be regarded as a novel suite of Business Intelligence (BI) methodologies. Unlike many BI tools that primarily excel in querying, reporting, and visualization, they often lack the sophistication required to fully harness process-mining capabilities (van der Aalst, 2011). [17]

#### 2.1.1 Brief history

The history of process mining can be traced back to the early 2000s when researchers began exploring methods to analyze event data generated by information systems [18]. One of the pioneering works in this field was the development of the Alpha algorithm by Wil van der Aalst and his colleagues in 2004 [19] which laid the foundation for automated process discovery. Subsequent years witnessed the emergence of various process-mining techniques aimed at extracting insights from event logs, including process conformance checking and performance analysis [20]. The field gained momentum with the introduction of commercial process mining tools in the late 2000s, facilitating its adoption across industries. As data availability increased and computational capabilities advanced, process mining evolved to address more complex process scenarios and integrate with other disciplines such as data science and artificial intelligence [21]. Today, process mining has become a vital component of digital transformation initiatives, enabling organizations to achieve greater transparency, efficiency, and agility in their operations.

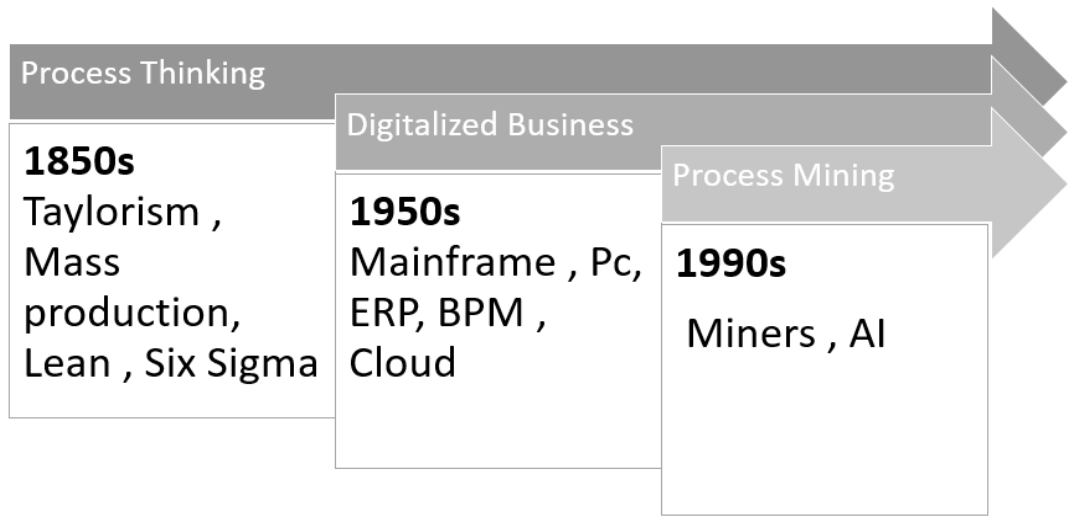

Figure 1: history of process mining

### 2.2. Process discovery

Process discovery is a fundamental aspect of process mining [22], focused on extracting and representing knowledge about business processes from event data recorded in information systems. The primary goal of process discovery is to create a clear and accurate representation of how activities are executed within an organization [23]. This involves identifying the sequence of tasks, decisions, and interactions that comprise a process, as well as the dependencies and relationships between them. Various techniques are employed for process discovery, ranging from straightforward methods such as alpha-algorithm and heuristic mining to more complex approaches like genetic algorithms and Petri net-based algorithms. Regardless of the technique used, process discovery plays a crucial role in gaining insights into process behavior, uncovering inefficiencies, and facilitating process improvement initiatives within organizations. By providing a visual representation of actual process executions, process discovery enables stakeholders to understand, analyze, and optimize business processes effectively.

#### 2.2.1 *Metrics for evaluating process models*

Metrics for evaluating process models play a vital role in assessing the quality and effectiveness of discovered process representations [24]. These metrics provide quantifiable measures that enable stakeholders to gauge various aspects of a process model's performance and suitability. Common metrics includes:

• **Fitness:** The fitness of a model indicates its ability to accurately replicate all sequences present in the event logs from start to finish. To measure fitness, the process model is executed against the event logs, and the degree of alignment between the model's behavior and the observed behavior in the logs is calculated. This alignment can be quantified using techniques such as the fitness function, which evaluates how well the model reproduces the sequences of activities present in the event logs.
• **Complexity:** Complexity is usually calculated based on the structural characteristics of the process model, such as the number of nodes (activities) and edges (transitions) in the model's graphical representation. The higher the number of nodes and edges, the greater the complexity of the model.

• **Precision:** Precision reflects the degree to which a model avoids "under-fitting." If the model permits additional behaviors beyond those evident in the current logs, it is considered under-fitted and lacks precision.
Precision is assessed by comparing the behavior captured by the model against the behavior observed in the event logs. If the model accurately represents the observed behavior without introducing additional, unobserved behaviors, it is considered precise. Techniques such as cross-validation or statistical analysis can be employed to evaluate precision.
• **Generalization:** Generalization pertains to the prevention of "overfitting." A model is considered insufficiently generalized if it solely explains the specific log sample, failing to produce a similar process model when applied to another sample log of the same process.

Overall, these metrics provide quantitative measures to assess different aspects of a process model's quality, helping organizations identify areas for improvement and make informed decisions about process optimization and refinement [25].

### 2.3. Process discovery algorithms

Process discovery algorithms are essential tools in process mining, aimed at automatically extracting and representing knowledge about business processes from event data. These algorithms analyze event logs generated by information systems to uncover the sequence of activities, decisions, and interactions that constitute a process. Various process discovery algorithms exist, each employing distinct techniques and approaches to infer process models from the observed data. Some common process discovery algorithms includes:

- Alpha Algorithm:
  • Utilizes a divide-and-conquer strategy to construct process models efficiently from event logs [26].
  • It breaks down the event log into smaller subsets and constructs local process models, which are then merged to form a global process model.
  • This algorithm is particularly effective for handling large event logs and is known for its scalability.

- Heuristic Mining:
  • Employs predefined rules and patterns to quickly discover process models.
  • It often relies on heuristics [27] to identify common patterns such as frequent sequences of activities, loops, or choices.
  • Heuristic mining is relatively straightforward and suitable for quickly obtaining insights from event data.

- Genetic Algorithms:
  • Mimic the process of natural selection to iteratively improve process model structures.
  • Genetic algorithms use a population of candidate solutions (process models) and evolve them over successive generations using techniques such as mutation, crossover, and selection [28].
  • They can effectively explore the search space of possible process models and converge towards optimal or near-optimal solutions.

- Petri Net-based Algorithms:
  • Leverage formal mathematical models to represent and analyze process behavior [29].
  • Petri nets provide a graphical and mathematical framework for modeling concurrent and asynchronous systems, making them well suited for capturing complex process behaviors.
  • Algorithms based on Petri nets translate event logs into Petri net models, which can then be analyzed and refined to represent the observed process behavior accurately.

These process discovery algorithms each have their strengths and weaknesses, and the choice of algorithm depends on factors such as the characteristics of the event data, the complexity of the process being analyzed, and the specific goals of the analysis.

## 3. PM4PY library

The PM4Py open-source library [34] serves as a robust and versatile toolkit for process mining tasks within the Python ecosystem. This section provides an overview of the PM4Py library, highlighting its key features, functionalities, and advantages for process mining implementation.

### 3.1. Features and Capabilities:

PM4Py offers a comprehensive set of features tailored specifically for process mining tasks, making it a valuable resource for researchers, practitioners, and data scientists. Some of the prominent features include:

• **Process Discovery**: PM4Py provides a range of algorithms for process discovery, enabling users to automatically extract process models from event log data. Algorithms such as the Alpha Miner, Inductive Miner, and Heuristic Miner offer different approaches to process discovery, catering to diverse process mining requirements.

• **Conformance Checking:** The library includes tools for assessing the conformance between discovered process models and observed behavior in event logs. Through conformance checking techniques, users can identify deviations, errors, and inefficiencies in process executions, facilitating process improvement initiatives.

• **Process Enhancement:** PM4Py facilitates process enhancement by offering functionalities for refining and optimizing process models based on discovered insights. Users can iteratively analyze, modify, and validate process models to achieve desired performance metrics and operational efficiencies.

• **Visualization and Interpretation:** PM4Py enables intuitive visualization of process models and analysis results, allowing users to interpret and communicate findings effectively. Visual representations aid in understanding process dynamics, identifying patterns, and communicating insights to stakeholders.

• **Method calling:** employs factory methods to implement its functionality. These methods serve as centralized entry points for each algorithm, offering a standardized set of input objects such as event data and parameter configurations. An example is the factory method for the Alpha Miner. Typically, these methods accept the algorithm variant's name along with certain parameters, which may be either common to all variants or specific to particular ones.

• **Object management:**

In process mining, the primary data source comprises event data, commonly known as event logs. These logs consist of a series of events that document the activities performed within different instances of the process being analyzed. PM4Py offers comprehensive support for various types of event data structures:

- Event Logs: These represent a collection of traces, where each trace is essentially a sequence of events. Each event is structured as a key-value map, providing detailed information about the activity and its attributes.

- Event Streams: These represent a single list of events, also organized as key-value maps. Unlike event logs, event streams are not yet organized into distinct cases or instances of processes.

PM4Py includes conversion utilities to seamlessly convert event data objects between these different formats. Additionally, the library facilitates the use of panda's data frames, which offer efficiency benefits when working with larger event data sets.

## 4. Case study on road traffic fine management process

In this demonstration, we aim to highlight the application of process mining techniques using an open-source library, applying process mining to a real-life event log that records the execution of steps in managing road traffic fines. The log includes the following steps : creating fines, sending fines, adding penalties, managing appeals, and handling payments. By leveraging this dataset, we will provide a systematic guide on extracting actionable insights from the fine management process data. Through process mining, we will systematically analyze the sequence of activities. This demonstration will offer users hands-on experience in employing process-mining methodologies to real-world scenarios. .

### 4.1. Importing data:

Importing data is a crucial initial step in any process mining endeavor, and one of the key components for this task is the "xes_importer" module within the pm4py library. By incorporating this module, accessed through the import statement shown in Fig2, users can efficiently import event logs stored in the XES (eXtensible Event Stream) format. The XES format is widely used for representing event data and is compatible with various process mining tools and platforms. Leveraging the capabilities of the xes_importer module, users can seamlessly read and load event log data into their Python environment, enabling subsequent analysis and exploration of process instances, activities, timestamps, and other relevant information. This streamlined import process lays the foundation for conducting comprehensive process mining analyses, empowering users to derive actionable insights and drive improvements in process performance and efficiency.

```
from pm4py.objects.log.importer.xes import importer as xes_importer
```

**Figure 2: importing a module from pm4py**

```
log[0]

{'attributes': {'concept:name': 'A1'}, 'events': [{'amount': 35.0,
'org:resource': '561', 'dismissal': 'NIL', 'concept:name': 'Create Fine',
'vehicleClass': 'A', 'totalPaymentAmount': 0.0, 'lifecycle:transition':
'complete', 'time:timestamp': datetime.datetime(2006, 7, 24, 0, 0,
tzinfo=datetime.timezone.utc), 'article': 157, 'points': 0}, '..',
{'concept:name': 'Send Fine', 'lifecycle:transition': 'complete', 'expense':
11.0, 'time:timestamp': datetime.datetime(2006, 12, 5, 0, 0,
tzinfo=datetime.timezone.utc)}]}
```

**Figure 3 : printing the first element of event log after importing the log**

When utilizing the pm4py.objects.log.importer.xes module to import an XES file log, the log is loaded into the program similar to a JSON object, allowing for structured access to its contents. Figure 3 demonstrates the process of accessing child objects within the loaded log. This means that users can navigate through the hierarchical structure of the log, accessing events, traces, attributes, and other relevant information as needed for analysis or further processing. By treating the imported log as an object with defined properties and methods, PM4Py provides a user-friendly interface for working with event data from XES files, enabling efficient exploration and manipulation of process mining data.

### 4.2. Analyzing data:

In the phase of analyzing the data, several crucial steps are undertaken to gain comprehensive insights into the process. One fundamental aspect involves identifying the starting and ending events within each trace of the event log. This entails examining the timestamps or attributes associated with events to pinpoint the initiation and conclusion of process instances. Additionally, the process mining analysis involves detecting variants within the event log, representing different paths or sequences of activities followed during process execution. By identifying and categorizing variants, analysts can uncover common patterns and deviations within the process flow. Moreover, the frequency of occurrence of each event is assessed to understand its significance and impact on process performance. This entails tallying the number of times each event occurs across all traces in the log. Finally, particular attention is given to determining the most prevalent variant, which signifies the pathway most frequently traversed by process instances. These analytical procedures enable process-mining practitioners to gain a holistic understanding of process dynamics and facilitate informed decision-making for process optimization and enhancement.

As shown in Figure 4 we have only one starting event

```
log_start = pm4py.stats.get_start_activities(log)
end_activities = pm4py.stats.get_end_activities(log)
log_start  # Printing the start activity in our log

{'Create Fine': 150370}
```

**Figure 4 : starting events**

Event logs in process mining may contain more than just a single start activity. In fact, they often reflect the dynamic and multifaceted nature of real-world processes by encompassing various initial activities that initiate different pathways within a process flow. These start activities represent the diverse entry points from which processes are initiated, capturing the complexity inherent in organizational workflows.

In the other hand as demonstrated in Figure 5, we have many different end activities.

```
end_activities

{'Send Fine': 20755,
 'Send for Credit Collection': 58997,
 'Payment': 67201,
 'Send Appeal to Prefecture': 3144,
 'Appeal to Judge': 134,
 'Notify Result Appeal to Offender': 86,
 'Receive Result Appeal from Prefecture': 53}
```

**Figure 5: end events**

Analyzing the distribution of start and end events within a process log offers valuable insights into process dynamics and efficiency. In this scenario, the most frequent start event, 'Create Fine', occurs 150,370 times, indicating a significant volume of instances where processes originate. On the other hand, end events such as 'Send Fine', 'Payment', and 'Send for Credit Collection' occur with varying frequencies, providing insights into the completion stages of different process pathways. Understanding the prevalence of start and end events can help identify bottlenecks, optimize process flows, and prioritize areas for improvement. For instance, a high frequency of 'Send for Credit Collection' might signal potential issues in revenue collection processes, prompting further investigation and refinement of related procedures to enhance overall efficiency and effectiveness.

### 4.3. Notion of variants:

A process variant represents a distinct pathway that traces the journey from the initial stage to the culmination of a process [30] . It encapsulates the specific sequence of activities, decisions, and events that occur within a process instance, providing a detailed narrative of its execution. Each process variant reflects a unique combination of steps taken, deviations encountered, and outcomes achieved throughout the process lifecycle. By analyzing process variants, organizations gain a granular understanding of the diverse ways in which processes unfold, allowing them to identify patterns, anomalies, and inefficiencies.
At first we can check how many variants we have in the log for the chosen dataset we have 231 variants as shown in Figure 6.

```
from pm4py.algo.filtering.log.variants import variants_filter

variants = variants_filter.get_variants(log)
print(f"We have:{len(variants)} variants in our log")

We have:231 variants in our log
```

**Figure 6 : showing how many variants in the log**

To deepen our understanding, we can extend our analysis by examining the first five variants, . By printing these initial variants, we gain valuable insights into the diverse paths that processes take within our system. This exploration allows us to grasp the variability and complexity inherent in our processes.

Interesting how Among the 150,370 cases logged, a substantial portion—specifically, 56,482 cases (approximately 2.8%)—belong to a single variant. Remarkably, this lone variant stands out among 231 variants documented. Upon closer examination, this variant reveals a concise sequence of five steps, culminating in what appears to be credit collection.
Furthermore, we can conduct an examination of the activities present in the log, taking into account their respective frequencies across all variants. This comprehensive analysis provides us with a holistic view of the activities undertaken within our processes, offering insights into their prevalence and distribution

### 4.4. Process discovery of the event log:

Having explored of the event log and gained a comprehensive understanding of its contents, the next step in our process mining endeavor involves the application of miner algorithms. As we embark on this phase, our objective is to explore various miner algorithms and assess their effectiveness in capturing the essence of the process. One critical aspect of this evaluation is comparing the simplicity and precision of the process models generated by each algorithm [25].

#### 4.4.1    *Simplicity: arc degree*

The simplicity evaluation framework assess the simplicity of a Petri net model by focusing on its degree, which represents the number of arcs within the model [31] . This metric suggests that a higher number of arcs might signify a higher level of complexity within the Petri net.

#### 4.4.2    *Precision: token based replay*

Token-based replay matches a trace and a Petri net model [32], starting from the initial place, in order to discover which transitions are executed and in which places we have remaining or missing tokens for the given process instance. Token-based replay is useful for Conformance Checking: indeed, a trace is fitting according to the model if, during its execution, the transitions can be fired without the need to insert any missing token. If the reaching of the final marking is imposed, then a trace is fitting if it reaches the final marking without any missing or remaining tokens.

### 4.5. Test results

In comparing different miners, the balance between simplicity and precision is crucial. While simpler models offer clarity and ease of interpretation, models that are more precise capture finer details of the process. Striking the right balance between these factors is essential for selecting an optimal process mining [33] approach tailored to the specific requirements and objectives of the analysis.

#### 4.5.1    *Example of result visualization*

Before we compare the metrics, let's see how managing road traffic fines twists and turns through a process discovery made possible by the alpha miner.

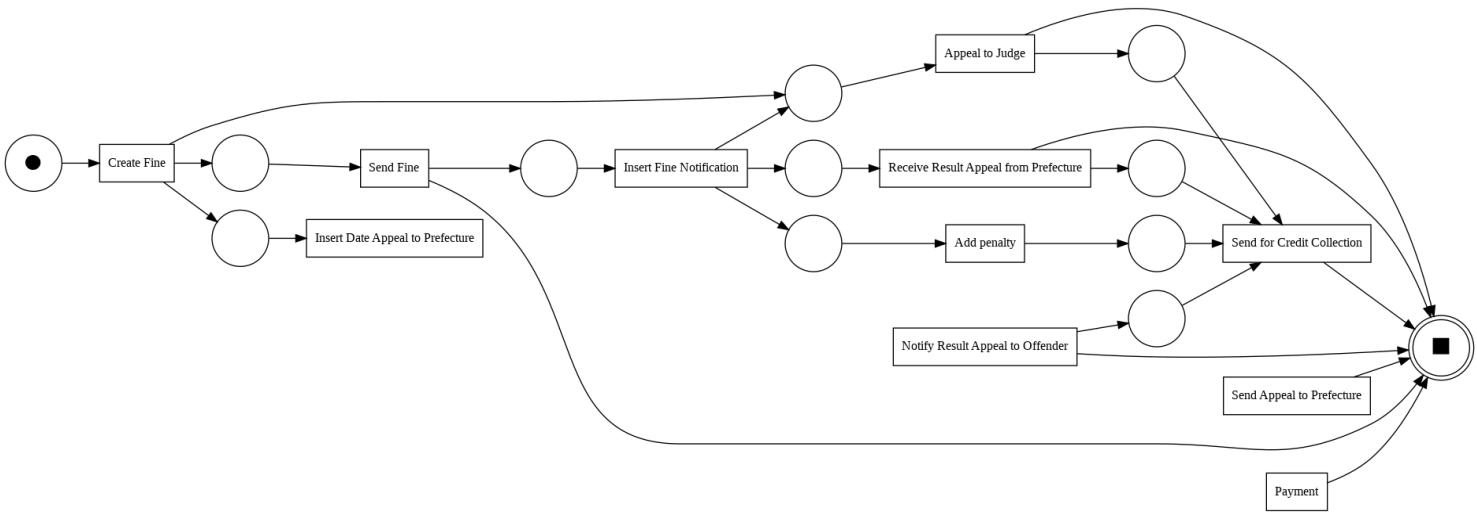

**Figure 7 : process discovery from alpha-miner**

The table below illustrates the varying levels of simplicity and precision exhibited by different mining algorithms.

**Table 1 : simplicity and precision of process miners**

| Miner | Simplicity | Precision |
| --- | --- | --- |
| Alpha | 0.66 | 0.66 |
| Alpha-plus | 0.45 | 0.66 |
| Inductive | 0.62 | 0.58 |
| Heuristic | 0.54 | 1.0 |

Upon comparing the simplicity and precision metrics across different miner algorithms, several observations emerge. The Alpha Miner algorithm exhibits moderate simplicity with a score of 0.66, closely followed by the Inductive Miner at 0.62. Interestingly, Alpha-plus, despite its name, presents lower simplicity at 0.45. In terms of precision, Alpha Miner, Alpha-plus, and Inductive Miner perform similarly with scores of 0.66, while Heuristic Miner stands out with a perfect precision score of 1.0. These findings underscore the importance of selecting a miner algorithm tailored to the specific trade-offs between simplicity and precision required for a given process mining task

## 5. Conclusion

In this paper, we conducted a comprehensive analysis of road traffic fine management processes utilizing the pm4py library in Python, aiming to understand the process dynamics and evaluate various process-mining algorithms. We started by importing an event log dataset and explored its characteristics, including activity distribution and process variants. Through filtering and statistical analysis, we uncovered key patterns and variations in process executions , at later stage we applied different process mining algorithms, such as the Alpha and alpha-plus miner , Inductive Miner, and Heuristic Miner, to discover process models from the event log data. Visualizing the discovered models allowed us to understand the workflow structures and dependencies within the process. Additionally, we discussed the strengths and limitations of each mining approach in capturing the underlying process dynamics.

The findings shed light on the efficiency and effectiveness of road traffic fine management processes, providing valuable insights for process optimization and decision-making. This study demonstrates the utility of pm4py in facilitating process mining tasks and its potential for analyzing real-world business processes.

Finally, we compared the simplicity and precision metrics across different miner algorithms. The Alpha Miner exhibited moderate simplicity, closely followed by the Inductive Miner, while Alpha-plus presented lower simplicity. In terms of precision, Alpha Miner, Alpha-plus, and Inductive Miner performed similarly, while Heuristic Miner stood out with a perfect precision score. These findings underscore the importance of selecting a miner algorithm tailored to the specific trade-offs between simplicity and precision required for a given process mining task.

In conclusion, this paper provides valuable insights into process mining methodologies, algorithms, and tools, with a focus on understanding the underlying mechanisms of process discovery.